# Finding the needle in high-dimensional haystack: A tutorial on canonical correlation analysis

Hao-Ting Wang[1,*], Jonathan Smallwood[1], Janaina Mourao-Miranda[2],
Cedric Huchuan Xia[3], Theodore D. Satterthwaite[3], Danielle S. Bassett[4,5,6,7],
Danilo Bzdok[,8,9,10,*]

[1]Department of Psychology, University of York, Heslington, York, YO10 5DD, United Kingdome
[2]Centre for Medical Image Computing, Department of Computer Science, University College London, London, United Kingdom; Max Planck University College London Centre for Computational Psychiatry and Ageing Research, University College London, London, United Kingdom
[3]Department of Psychiatry, Perelman School of Medicine, University of Pennsylvania, Philadelphia, PA, 19104, USA
[4]Department of Bioengineering, University of Pennsylvania, Philadelphia, Pennsylvania 19104, USA
[5]Department of Electrical and Systems Engineering, University of Pennsylvania, Philadelphia, Pennsylvania 19104, USA
[6]Department of Neurology, Perelman School of Medicine, University of Pennsylvania, Philadelphia, PA 19104 USA
[7]Department of Physics & Astronomy, School of Arts & Sciences, University of Pennsylvania, Philadelphia, PA 19104 USA
[8] Department of Psychiatry, Psychotherapy and Psychosomatics, RWTH Aachen University, Germany
[9]JARA-BRAIN, Jülich-Aachen Research Alliance, Germany
[10]Parietal Team, INRIA, Neurospin, Bat 145, CEA Saclay, 91191, Gif-sur-Yvette, France

* Correspondence should be addressed to:
Hao-Ting Wang
haoting.wang@york.ac.uk
Department of Psychology, University of York, Heslington, York, YO10 5DD, United Kingdome

Danilo Bzdok
danilo.bzdok@rwth-aachen.de
Department of Psychiatry, Psychotherapy and Psychosomatics, RWTH Aachen University, Germany

# 1 ABSTRACT

Since the beginning of the 21$^{st}$ century, the size, breadth, and granularity of data in biology and medicine has grown rapidly. In the example of neuroscience, studies with thousands of subjects are becoming more common, which provide extensive phenotyping on the behavioral, neural, and genomic level with hundreds of variables. The complexity of such "big data" repositories offer new opportunities and pose new challenges to investigate brain, cognition, and disease. Canonical correlation analysis (CCA) is a prototypical family of methods for wrestling with and harvesting insight from such rich datasets. This doubly-multivariate tool can simultaneously consider two variable sets from different modalities to uncover essential hidden associations. Our primer discusses the rationale, promises, and pitfalls of CCA in biomedicine.

**Keywords**: machine learning, data science, modality fusion, deep phenotyping

## 2 MOTIVATION

The combination of parallel developments of large biomedical datasets and increasing computational power have opened new avenues with which to understand relationships among brain, cognition, and disease. Similar to the advent of microarrays in genetics, brain-imaging and extensive behavioral phenotyping yield datasets with tens of thousands of variables (1). Since the beginning of the 21$^{st}$ century, the improvements and availability of technologies, such as functional magnetic resonance imaging (fMRI), have made it more feasible to collect large neuroscience datasets (2). At the same time, problems in reproducing the results of key studies in neuroscience and psychology have highlighted the importance of these large datasets (3).

For instance, the UK Biobank is a prospective population study with 500,000 participants and comprehensive imaging data, genetic information, and environmental measures on mental disorders and other diseases (4,5). Similarly, the Human Connectome Project (6) has recently completed brain-imaging of >1,000 young adults, with much improved spatial and temporal resolution, with overall four hours of body scanning per participant. Further, the Enhanced Nathan Kline Institute Rockland Sample (7) and the Cambridge Centre for Aging and Neuroscience (8,9) offer cross-sectional studies (n > 700) across the lifespan (18–87 years of age) in large population samples. By providing rich datasets that include measures of brain imaging, cognitive experiments, demographics, and neuropsychological assessments, such studies can help quantify developmental trajectories in cognition as well as brain structure and function. While "deep" phenotyping and large sample sizes provide opportunities for more robust descriptions of subtle population variability, the data abundance does not come without new challenges.

Some classical statistical tools may struggle to fully profit from datasets that provide more variables than observations of these variable sets (10–12). Even large samples of participants are smaller in number than the number of brain locations that have been sampled in high-resolution brain scans. On the other hand, in datasets with a particularly high number of participants, traditional statistical approaches will identify associations that are highly statistically significant but only account for a small fraction of the variance (5,12). In the "big data" setting, investigators can hence profit from extending the traditional toolbox of under-exploited quantitative analysis methods.

The present tutorial advocates canonical correlation analysis (CCA) as a tool for charting and generating understanding from modern datasets. One key property of CCA is that this tool can simultaneously evaluate two different sets of variables. For example, CCA allows a data matrix comprised of brain measurements (e.g., connectivity links between a set of brain regions) to be analyzed with respect to a second data matrix comprised of behavioral measurements (e.g., response items from various questionnaires). CCA simultaneously identifies the sources of variation that bear strongest statistical associations between both sources of variation.

More broadly, CCA is a multivariate statistical method that was introduced in the 1930ies (13). However, besides being data-hungry, CCA is computationally expensive and has therefore only become increasingly applied in biomedical research relatively recently. Moreover, the ability to accommodate two multivariate variable sets allows the identification of patterns that describe many-to-many relations. CCA provides utility that goes beyond techniques that map one-to-one relations (e.g., Pearson's correlation) or many-to-one relationships (e.g., ordinary multiple regression). With the emergence of larger datasets, researchers in neuroscience, and other biomedical sciences, have recently begun to take advantage of these qualities of CCA to ask and answer novel questions regarding the links between brain, cognition, and disease (14–19).

Our tutorial provides investigators with a road map for how CCA can be used to to most profitably understand questions in fields such as cognitive neuroscience that depend on uncovering patterns in complex datasets. We first introduce the computational model and the circumstances of use with recent applications of CCA in existing research. Next, we consider the types of conclusions that can be drawn from application of the CCA algorithm, directing special attention to the limitations of this technique. Finally, we provide a set of practical guidelines for the usage of CCA in future scientific investigations.

## 3 MODELING INTUITIONS

One way to appreciate the modeling goal behind CCA is by viewing this procedure as an extension of principal component analysis (PCA). This widespread matrix decomposition technique identifies a set of latent dimensions as a linear approximation of the main variance components that underlie the information dispersed across the original variables. In other words, PCA can re-express a set of correlated variables in a smaller number of hidden factors of variation. These latent sources of variability are not directly observable in the original data, but scale and combine to collectively explain how the actual observations may have come about as component mixtures.

PCA and similar matrix-decomposition approaches have been used frequently in the domain of personality research. For example, the 'Big Five' describes a set of personality traits that are identified by latent patterns that are revealed when PCA is applied to the answers people give to personality tests (20). This approach tends to produce five reliable components that explained a substantial amount of meaningful variation in data gathered by personality tests. A strength of a decomposition method such as PCA is the ability for parsimonious reduction of the original datasets into continuous dimensional representations. These can also often be amenable to human interpretation (such as the concept of introversion). The ability to re-express the original data in a compressed, more compact form has computational, statistical, and interpretational appeal.

In a similar vein to PCA, CCA maximizes the linear correspondence between two parallely decomposed sets of variables. The algorithm seeks dimensions that describe shared variance across both sets of measures. In this way, CCA is particularly useful when describing observations that bridge two levels of observation with the aim of modality fusion. Examples include i) genetics and behavior, ii) brain and behavior, or iii) brain and genetics. Three central properties underlying data modeling using CCA are joint-information compression, symmetry and multiplicity, which we discuss next.

### 3.1 JOINT INFORMATION COMPRESSION

The purpose of CCA is to find linked sources of variability that describe the characteristic correspondence between two sets of variables, typically capturing two different levels of observation. The relations between each pair of factors across variable sets can be used to compute notions of the conjoined variance explained across both domains. Similar to PCA, CCA aims to find useful projections of the high-dimensional variable sets onto the compact linear representations, the *canonical variates*. Each resulting canonical variate is computed from the weighted sum of every original variable indicated by the *canonical vector*. Yet, the information compression goal is based on maximizing the linear correspondence between the low-rank projections from each side under the constraint of uncorrelated hidden dimensions, often referred to formally as orthogonality. Again similar to PCA, these dimensions are naturally ranked by their explained variance, with first dimensions accounting for most variability in the data. The *canonical correlation* quantifies the linear correspondence between the left and right variables sets based on Pearson's correlation between their canonical variates; how much the right and left variable set can be approached to each other in the embedding space (Fig 1). Canonical correlation can be seen as a metric of successful joint information reduction between two variable arrays and, therefore, routinely serves as a performance measure for CCA. As a logical and practical consequence of its doubly multivariate nature, adding or removing even a single variable in one of the variable sets can lead to larger changes in the CCA solution (21). Thus, care must be taken in assessing reliability and robustness of results. To summarize the notion of joint information compression: CCA finds any linear combination of a set of variables that most highly correlates with any linear combination of another set of variables.

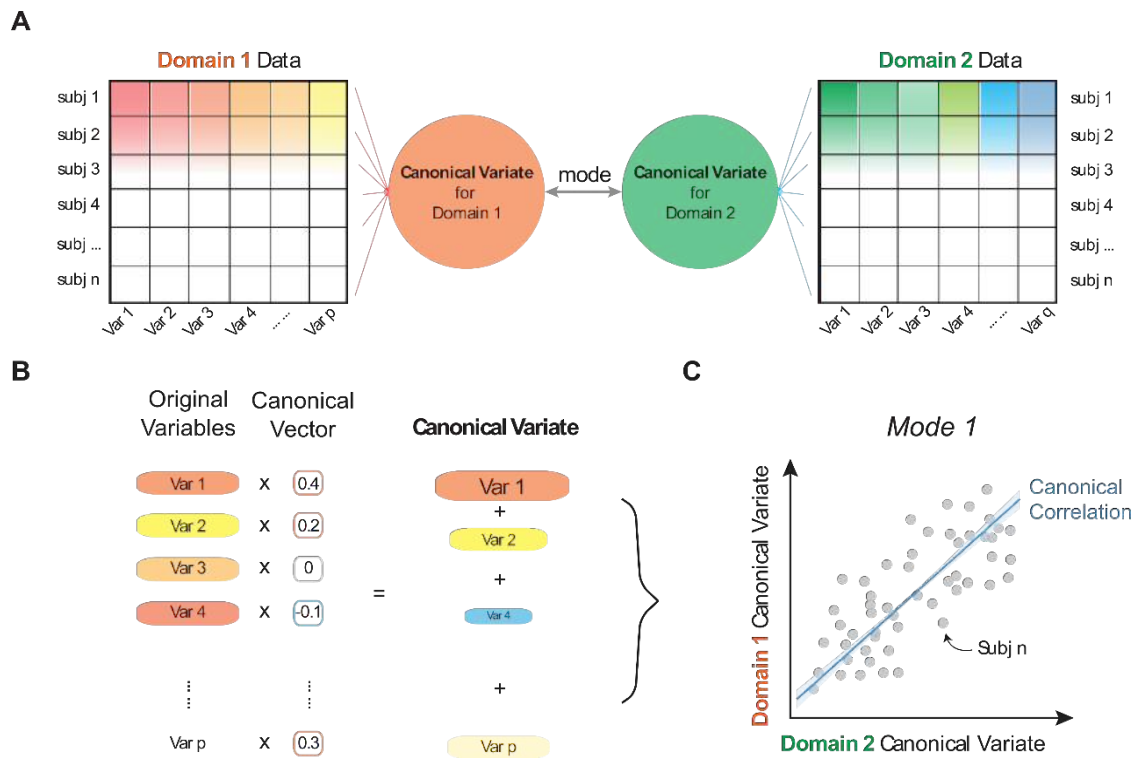

**Fig 1. A general schematic for Canonical Correlation Analysis (CCA).**
(A) Multiple domains of data, with *p* and *q* variables respectively, measured on the same sample can be submitted to co-decomposition by CCA. The algorithm seeks to re-express the datasets as multiple pairs of canonical variates that are highly correlated with each other across subjects. Each pair of the latent embedding of the left and right variable set can be called a mode. (B) In each domain of data, the resulting canonical variate is composed of the weighted sum of variables by the canonical vector. (C) In a two-way CCA setting, each subject can thus be parsimoniously described two canonical variates per mode, which are maximally correlated as represented here on the scatter plot. The linear correspondence between these two canonical variates is the canonical correlation - a primary performance metric used in CCA modeling.

## 3.2 SYMMETRY

One important feature of CCA is that the two variable sets are mutually exchangeable. Swapping the left and right sides yields identical estimates of canonical variates and canonical correlations. Put differently, the classical notions of 'independent variables' or 'explanatory variables' usually denoting the model input (e.g., several questionnaire response items) and 'dependent variable' or 'response variable' usually denoting the model output (e.g., total working memory performance) lose their meaning in the context of CCA (22). The canonical correlation remains identical because the underlying linear correlation is a symmetric bivariate metric: A unit change in one series of measurements is related to another series of measurements in another set of observations. The first set of values thus depends in the same way on the second set of values as the second set of values depends on the first.

The notion of symmetry is a key property that distinguishes CCA from many other linear-regression methods, specifically those where dependent and independent variables play operationally distinct roles during model estimation. For instance, linear regression accounts for the impact of a unit change in the (dependent) response variable as a function of the (independent) input variable. Thus the dependent and independent variables cannot be exchanged to obtain an identical result. In short, as CCA estimation relies on optimizing a correlation-based performance metric, this method describes the co-relationship between two sets of variables regardless of which set is on the left or on the right side of the equation.

## 3.3 Multiplicity

In CCA, a particular pair of canonical variates that jointly share variance in both variable sets is known as a *mode*. Each CCA mode can thus be described by one low-rank projection of the left variable set (one canonical variate associated with that mode) and another low-rank projection of the right variables second (the other canonical variate associated with that mode). Intuitively, after extracting the first and most important mode, indicated by the highest canonical correlation, CCA can determine the next pair of latent dimensions (i.e., canonical vectors) whose variance between both variable sets has not yet been explained by the first mode. Since every new mode is found in the residual variance in the variable arrays, so in the classical formulation of CCA the modes are optimized to be mutually uncorrelated with each other (i.e., orthogonality constraint). As such, analogous to PCA, CCA produces a set of mutually uncorrelated modes naturally ranked by explained variance. The orthogonality constraint imposed during CCA estimation ensures that the modes represent unique variability patterns that describe different aspects of the data. Technically, the collection of all modes can also be obtained by computing the singular value decomposition of the dot product of both variable sets (23). When the modes turn out to be scientifically meaningful, one attractive interpretation can revolve around overlapping descriptions of processes of behavior, brain, gene expression. For instance, much genetic variability in Europe can be jointly explained by a north-south axis (i.e., one mode of variation) and a west-east axis (i.e., another mode of variation) (24).

Figure 1 illustrates how the three core properties underlying CCA modeling make it a particularly useful technique for the analysis of modern biomedical datasets – joint information, compression and multiplicity. First, CCA can provide an effective hidden representation that succinctly captures the variance present in the original variables. Next, the CCA model is symmetrical in the sense that no numerical difference happens in the exchange of the two variable sets. Finally, we can estimate a collection of modes that quantitatively describe the correspondence between any two variable sets. As such, the purpose of CCA modeling departs from the common focus on "true" effects of single variables and instead targets prominent correlation structure shared across dozens or potentially thousands of variables (25). Together these allow CCA to efficiently uncover symmetric linear relations that compactly summarize doubly-multivariate data.

## 3.4 EXAMPLES

In recent studies in neuroscience, but also genomics, transcriptomics and many other data-rich empirical fields, we see an increasing number of CCA applications. Smith and colleagues (15) leveraged CCA to uncover brain-behavior modes of population co-variation in approximately 500 healthy participants from the Human Connectome Project (6). These investigators aimed to discover whether specific patterns of whole-brain functional connectivity, on the one hand, are associated with specific sets of various demographics and behaviors on the other hand (see Fig 2 for the analysis pipeline). Functional brain connectivity was estimated from resting state functional MRI scans measuring brain activity in the absence of a task or stimulus (26). Independent component analysis (ICA ,27) was used to extract 200 network nodes from fluctuations in neural activity fluctuations. Next, functional connectivity matrices were calculated based on the pairwise correlation of the 200 nodes to yield a first variable set that quantified inter-individual variability in brain connectivity "fingerprints" (28). A rich set of personal measures ranging from cognitive performance indices to demographic profiles provided a second variable set that captured inter-individual variability in behavior. The two variable arrays were submitted to CCA to gain insight into how latent dimensions of network coupling patterns present linear correspondences to latent dimensions underlying phenotypes of cognitive processing and life experience. The statistical robustness of the ensuing brain-behavior modes was determined via a non-parametric permutation approach using the canonical correlation as the test statistic. A single statistically significant CCA mode emerged, which exhibited strong population-level co-variation of whole-brain network connectivity measures and diverse behavioral measures. This most important CCA mode was comprised of behavioral measures that varied along a positive-negative axis; measures of intelligence, memory, and cognition were located on the positive end of the mode, and measures of lifestyle were located on the negative end of the mode. The brain regions exhibiting strongest contributions to coherent connectivity changes were reminiscent of the default mode network (29). It is notable that prior work has provided evidence that regions composing the default mode network are associated with episodic and semantic memory, scene construction, and complex social reasoning such as theory of mind (30–32). The positive-negative dimensions in the behavioral component and the modulated functional couplings related to the default mode network were formally discovered and quantitatively described in a single CCA model fit. The finding of Smith and colleagues (15) provide evidence that functional connectivity in the default mode network is important for higher-level cognition and intelligent behaviors and that have important links to life satisfaction.

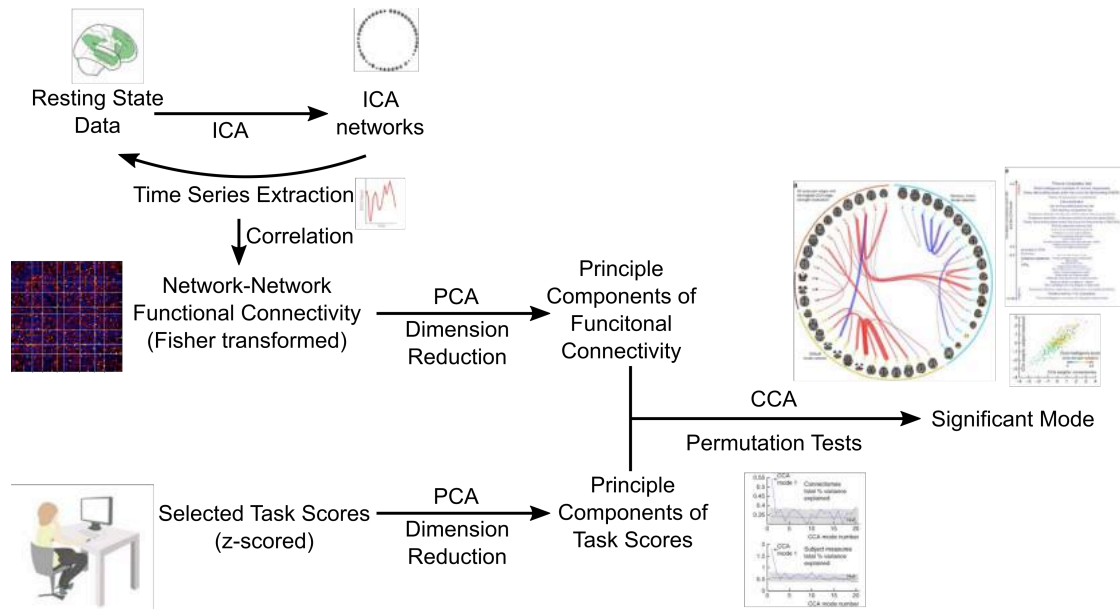

**Fig 2. The analysis pipeline of Smith et al., 2015.**
These investigators aimed to discover whether specific patterns of whole-brain functional connectivity, on the one hand, are associated with specific sets of correlated demographics and behaviors on the other hand. The two domains of the input variables were transformed into principle components before the CCA model evaluation. The significant mode was determined by permutation tests. The finding of Smith and colleagues (2015) provide evidence that functional connectivity in the default mode network is important for higher-level cognition and intelligent behaviors and are closely linked to positive life satisfaction.

Another use of CCA has been to help understand the complex relationship between neural function and patterns of ongoing 'spontaneous' thoughts. In both the laboratory and in daily life, ongoing thought can often shift from the task at hand, to other personally relevant characteristics - a phenomenon that is often referred to by the term 'mind-wandering' (34,35). These shifts of attention from the immediate environment to constructed internal representations have been associated with poorer performance on attention-demanding tasks (36,37). However, other studies focused on problem-solving suggest that mind wandering may promote creativity (38,39) and future planning (38,40). This heterogeneity of functional outcomes led to the hypothesis that different patterns of ongoing thought may have unique functional associations, thus explaining why mind-wandering has such a complex set of psychological correlates. Wang and colleagues (33) used CCA to empirically explore this question by examining the links between connectivity within the default mode network and patterns of ongoing self-generated thought recorded in the lab (Fig 3). Their analysis used patterns of functional connectivity within the default mode network as one set of observations, and self-reported descriptions recorded in the laboratory across multiple days as the second set of observations (21). The connectivity among 16 regions in the default mode network and 13 self-reported aspects on mind-wandering experience were fed into a sparse version of CCA (see below for different CCA variants). This analysis found two stable patterns that were termed as positive-habitual thoughts and spontaneous task-unrelated thoughts, both associated with unique patterns of connectivity fluctuations within the default mode network. As a means to further validate the extracted brain-behavior modes in new data (11), follow-up analyses confirmed that the modes were uniquely related to aspects of cognition, such as executive control and the ability to generate information in a creative fashion, and the modes also independently distinguished well-being measures. These data suggest that the default mode network can contribute to ongoing thought in multiple ways, each with unique behavioral associations and underlying neural activity combinations. By demonstrating evidence for multiple brain-experience relationships within the default mode network, Wang and colleagues (33), underline that greater specificity is need when considering the links between brain activity and neural experience (see also ,34).

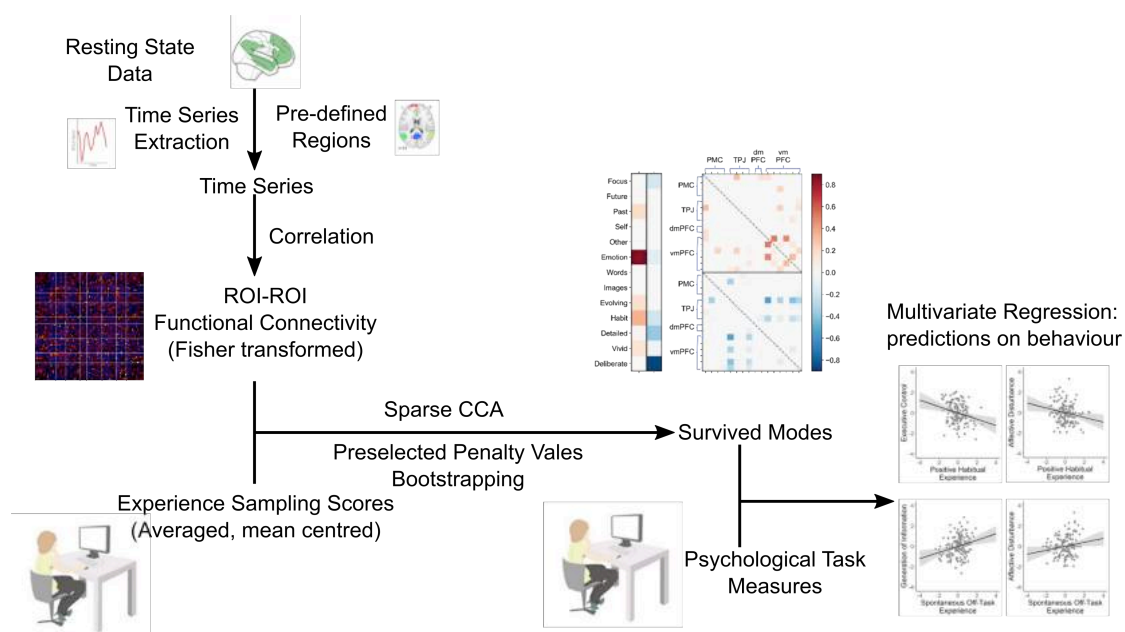

**Fig 3. The analysis pipeline of Wang et al., 2018b.**
Wang and colleagues (2018b) used CCA to interrogate the hypothesis that various distinct aspects of ongoing thought can track distinct components of functional connectivity patterns within the default mode network. Sparse CCA was used to perform feature selection simultaneously with the model fitting on the brain-experience data. The identified CCA modes showed robust trait combinations of positive-habitual thoughts and spontaneous task-unrelated thoughts with linked patterns of connectivity fluctuations within the default mode network. The two modes were also related to distinct high-level cognitive profiles respectively. These findings substantiate emerging evidence that different functional configurations of the default mode network may contribute to its integrative role in global network reconfiguration and in advanced cognition beyond its initially posited role in task-unrelated house-keeping functions.

In the third and final example, Xia and colleagues (14, see Fig 4) mapped item-level psychiatric symptoms to brain connectivity patterns in brain networks using resting-state fMRI scans in a sample of 1000 subjects from the Philadelphia Neurodevelopmental Cohort (44). Recognizing the marked level of heterogeneity and comorbidity in existing diagnostic psychiatric diagnoses, these investigators were interested in how functional connectivity and individual symptoms can form linked dimensions of psychopathology and brain networks (45). Notably, the study used a feature-selection step based on median absolute deviation to first reduce the dimensionality of the connectivity feature space prior to running CCA. As a result, about 3000 functional edges and 111 symptom items were jointly analyzed. As the number of features was still greater than the number of subjects, sparse CCA was used (21), which is a variant in the CCA family that penalizes the number of features selected by the final CCA model. Based on covariance-explained and subsequent permutation testing (46), the analysis identified four linked dimensions of psychopathology and functional brain connectivity – mood, psychosis, fear, and externalizing behavior. Through a resampling procedure that conducted sparse CCA in different subsets of the data, the study identified stable clinical and connectivity features that consistently contributed to each of the four modes. The resultant dimensions were relatively consistent with existing clinical diagnoses, but additionally cut across diagnostic boundaries to a significant degree. Furthermore, each of these dimensions were associated with a unique pattern of abnormal connectivity. However, the loss of network segregation was common to all dimensions, particularly between executive networks and the default mode network. As network segregation is a normative feature of network development, loss of network segregation across all dimensions suggests that common neurodevelopmental abnormalities may be important for a wide range of psychiatric symptoms (47). Taking advantage of CCA's ability to capture common sources of variation in more than one datasets, these findings support the idea behind NIMH Research Domain Criteria that specific circuit-level abnormalities in the brain's functional network architecture may give rise to a diverse psychiatric symptoms (48).

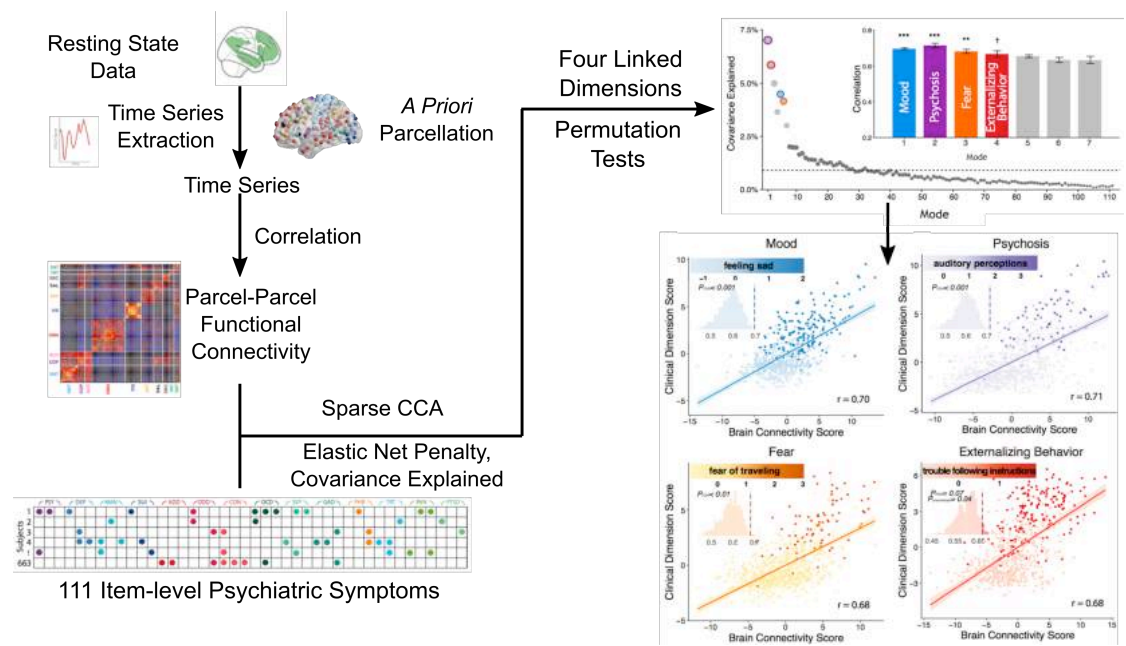

**Fig 4. The analysis pipeline of Xia et al., 2018.**
Xia and colleagues (2018) were interested in how functional connectivity and individual symptoms can form linked dimensions of psychopathology and brain networks. The study took a feature selection step based on median absolute deviation in preprocessing to first reduce the dimensionality of the functional connectivity measures. A sparse variation of CCA was applied to extract modes of linked dimensions of psychopathology and functional brain connectivity. Based on covariance-explained and subsequent permutation testing, the analysis identified four linked dimensions – mood, psychosis, fear, and externalizing behavior – each were associated with a unique pattern of abnormal brain connectivity. The results suggested that specific circuit-level abnormalities in the brain's functional network architecture may give rise to a diverse psychiatric symptoms.

# 4 INTERPRETATION

The conjoint matrix decomposition goal of CCA makes this modeling tool particularly useful for getting a handle on richly sampled descriptions of studied phenomenon from different levels of investigation. Yet, it is a matter of ongoing debate whether this analysis technique corresponds more closely to a *descriptive* re-expression of the data (i.e., unsupervised modeling) or should be more readily understood as a form of *predictive* reduced-rank regression (21,25,i.e., supervised modeling, cf. ,49). There are legitimate arguments in support of both views. A supervised algorithm depends on a designated modeling target to be predicted from an array of input variables, whereas an unsupervised algorithm aims to extract coherent patterns in observations without associated ground-truth labels that can be used during model estimation (10,23). As one possible intuition, as the dimensionality of one of the variable sets declines to approach the single output of most linear-regression-type methods, the more CCA applications resemble a supervised modeling approach. Conversely, with increasingly large variable sets on both sides, applying CCA is perhaps closer in spirit to an unsupervised modeling approach.

Whether the investigator considers CCA as either a supervised or unsupervised method has critical consequences for interpretation and for choosing eligible strategies to validate the model solutions. Cross-validation is a commonly seen technique for supervised model evaluation by comparing model-derived predictions in unseen data, but seldom used to buttress unsupervised model solutions, such as clustering methods like k-means or matrix decomposition techniques like PCA, due to the lack of labels for performance estimations (10,23,50). In an unsupervised setting, there is typically no unambiguously best optimality criterion (such as low residual sum of squares in supervised linear regression) that could be used for model selection or model evaluation, such as in cross-validation schemes (10,23). In a supervised setting, cross-validation procedures provide an accurate estimate of a model's capacity to generalize to future data samples. CCA is a somewhat unique case in this regard. A CCA model describes the data without an existing ground truth, cross-validation procedures evaluate CCA model by projecting data from new, previously unseen individuals using previous obtained canonical vectors to estimate the canonical correlation strength in new data, rather than accuracy of specific variable-variable links. One alternative validation strategy is to demonstrate that the canonical variates of the CCA fit are useful in explaining variance in other measurements of differences between (already seen) individuals (e.g. ,19). Another alternative validation strategy is to show that the CCA solutions are robust when repeating the analysis on random subsets of the (already seen) individuals in split-half analyses (5,15).

Furthermore, from a more formal perspective, the optimization objective governing parameter estimation during CCA fitting is unusual for a supervised model by being based on Pearson's correlation. The majority of linear-regression-type predictive models instead have an optimization loss that accumulates the extent of deviation from the ground-truth labels, including different residual-sum-of-squares loss functions (23,51). Moreover, as mentioned above, symmetry of the variable sets can be taken as another argument that pushes CCA more towards the unsupervised side of the model space. We are unaware of many supervised predictive models that yield identical sets of model parameter fits after the independent and dependent variables have been swapped. In conclusion, it is our impression that the CCA model is a special "animal" that sits in between the classical distinction between supervised and unsupervised methodology.

More broadly, statistical methods have been proposed to primarily fit into one of three categories based on their modeling goal: *estimation*, *prediction*, or *inference* (52,53). Model estimation typically refers to the raw process of adjusting randomly initialized parameters by fitting them to the data at hand; an intuitive example of such parameters are the beta parameters in linear regression to adjust a line to data points. As model estimation is often performed without necessarily applying the model to unseen observations or assessing the fundamental trueness of the effects, some authors recently called this modeling regime "retrodiction" (54,55). Prediction in turn is concerned with maximizing model fit to still be optimally useful in previously unseen data that would only be observed in the future. Drawing inference on model fits has frequently been based on statistical null hypothesis testing and accompanying methodology (56). This form of drawing rigorous conclusions from data is especially useful in classical settings to make precise statements about the contribution of single input variables.

Given this previously proposed general distinction of modeling goals, CCA probably most naturally qualifies for the estimation category, rather than qualifying as a primarily predictive or inferential tool. Because of its exploratory character of CCA, it is useful for applications to focus on uncovering parsimonious structure in two high-dimensional spaces as alternative descriptions of the observations at hand. Identifying predictive value of individual variables in new data is not an integral part of the optimization objective underlying CCA. Typical CCA applications do not necessarily seek to establish statistically significant links between particular *subsets* of the many variables of the two sets, since the goal is targeted at relevant patterns found across the entirety of two variable sets. Even if p-values are obtained based on non-parametric null hypothesis testing in the context of CCA, the particular null hypothesis at play is really centered on the *overall* robustness of the latent space correlations, as measured by the canonical correlations between the (projected) variable sets, and is not centered on specific variable-variable links. Thus, using CCA to pinpoint specific relations warrants sober caution, and care in interpreting the model findings. Stated in yet another way, CCA is typically a suboptimal choice when the investigator wishes to make strong statements about the relevance and relationships of individual variables of a variable set - a property shared with many other machine-learning tools.

### 4.1 LIMITATIONS OF CCA

We now discuss several challenges that researchers may encounter when considering whether CCA is a good choice for a given data-analysis problem. We summarize several important choices in the form of a flowchart (see Fig 5). As with many statistical approaches, the number of observations *n* in relation to the number of variables *p* is a key aspect when considering whether CCA is a promising choice (57,58). Ordinary CCA can only be expected to yield useful model fits in data with more observations than the number of variables of the larger variable set (i.e., $n > max(p, q)$). Concretely, if the number of individuals included in the analysis is too close to the number of brain or behavior or genomics variables, then CCA will struggle to well approximate the latent dimensions in the population (but see regularized CCA variants below). Concretely, even when the CCA reaches a solution, without throwing an error, the derived canonical vectors can be meaningless (58). More formally, in such degenerate cases, CCA loses its usual quality to find unique identifiable solutions (despite being a non-convex optimization problem) that another laboratory with the same data and CCA implementation could also obtain (59). Additionally, with increasing number of variables in one or both of the sets, the ensuing canonical correlation often tends to increase, due to higher degrees of freedom. An importance consequence is that canonical correlation obtained from CCA applications with differently sized variables sets cannot be used to decide which the obtained CCA models should be selected as "better". The CCA solution is constraint by the sample as well as the number of features. As a cautionary note, the canonical correlation effect sizes obtained from the training data limit statements about how that CCA solution would perform on future or other data.

In a similar vein, smaller datasets offering measurements from only a few dozen observations may have difficulty in fully profiting from the strengths of a flexible doubly multivariate procedure such as CCA. The ground-truth effects in areas like psychology, neuroscience, and genetics are often expected to be small, which are even harder to detect with insufficient sampling of the variability components. One practical remedy that can alleviate modeling challenges in small datasets is using data reduction methods such as PCA or other data-reduction method for preprocessing of each variable set performed before applying CCA (e.g. ,15) or adopt a sparse variations (see below). Reducing the variable sets according to their most important directions of linear variation can facilitate the doubly-multivariate search, while the ensuing CCA solution, including canonical variates, can be translated back to and interpreted within the original variable space. These considerations illustrate why CCA applications have long been less attractive in the context of many neuroscience studies, while its appeal and feasibility are now steadily growing as evermore rich, multi-modal, and open datasets become available (2,60).

A second limitation concerns the scope of the statistical relationships that CCA can discover and quantify in the underlying data. As a linear model, classical CCA imposes the assumption of additivity on the underlying relationships to unearth relevant linked co-variation patterns, thus ignoring more complicated variable-to-variable relationships that may exist in the data. CCA can accommodate any metric variable without strict dependence on normality. However, Gaussian normality in the data is desirable because covariance-based CCA exactly operates on differences in means and variances that parameterize this data distribution. Before CCA is applied to the data, it is common practice that one evaluates the normality of the variable sets and possibly apply data an appropriate transformation, such as z-scoring (variable normalization by mean centering to zero and unit-variance scaling to one) or Box-Cox transformations (variable normalization based on logarithm and square-root operations). Finally, the relationships discovered by CCA solutions have been optimized to highlight those variables whose low-dimensional projection is most (linearly) coupled with the low-dimensional projection of the other variable set. As such, the derived modes provide only a window into which multivariate relationships are most important given presence of the *other variable set,* rather than identifying variable subsets that are important in the dataset per se.

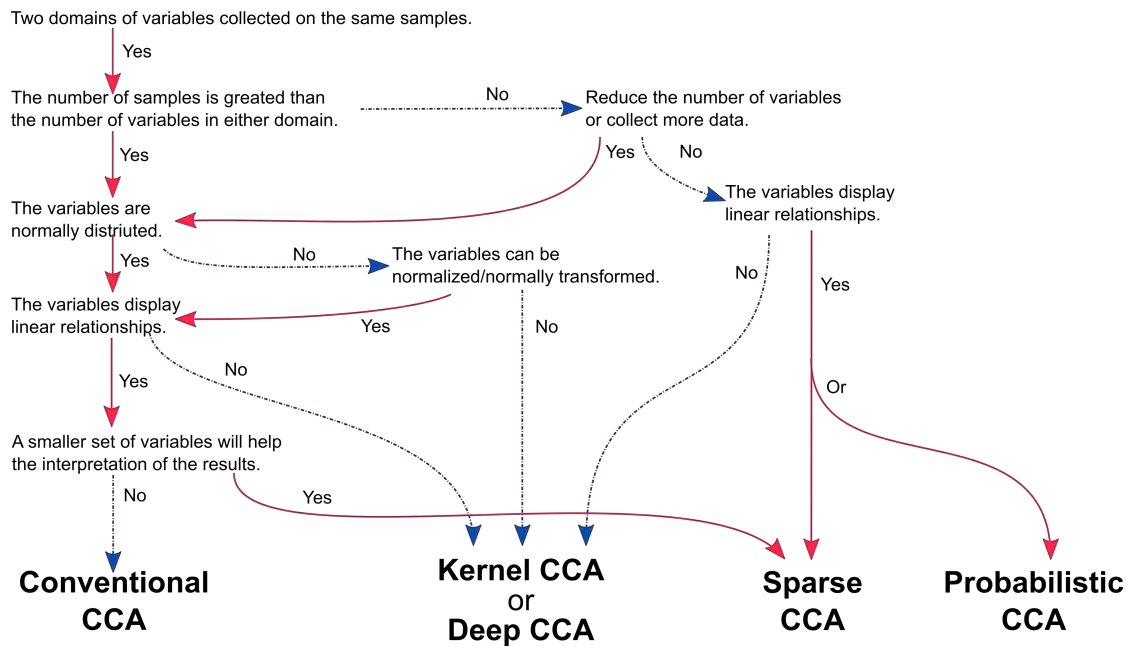

**Fig 5. A flowchart illustrating the choices when considering the application of CCA a dataset.**
The most important criteria is to have a dataset with two domains of multivariate variables. This flowchart serves as an advice rather than strict guidance. The choice of CCA variation can be determined by interpretation preferences (i.e. conventional vs sparse CCA and sparse CCA vs probabilistic CCA.).

## 4.2 RELATION TO OTHER METHODS AND CCA EXTENSIONS

CCA is probably the most general statistical approach to distill the relationships between two high-dimensional sources of quantitative measurements. CCA can be viewed as a broad class of methods that generalizes many more specialized approaches from the general linear model (GLM ,61). In fact, most of the linear models in common use by behavioral scientists for parametric testing, including ANOVA, MANOVA, multiple regression, Pearson's correlation, and t-test, can be interpreted as special cases of CCA (62,63). Because these techniques bear intimate relations, getting a grasp of the opportunities and challenges of various CCA approaches can help conceptual understanding of a large variety of modeling approaches throughout applications in biomedicine.

Related methods:
i) **PCA** has certain similarities to CCA, although PCA performs unsupervised matrix decomposition of one variable set (64). A shared property of PCA and CCA relies in the orthogonality constraint imposed during structure discovery. As such, the set of uncovered sources of variation (i.e., modes) are assumed to be uncorrelated with each other in both methods. As an important difference, there are PCA formulations that minimize the reconstruction error between the original variable set and the back-projection of each observation from the latent dimensions of variation (58). CCA instead directly optimizes the correspondence between the latent dimensions directly in the embedding space, rather than the reconstruction loss in the original variables incurred by the low-rank bottleneck. Moreover, PCA can be used for dimensionality reduction as a pre-processing step before CCA (e.g. ,15).

ii) Analogous to PCA and CCA, **independent component analysis (ICA)** also extracts hidden dimensions of variation in a potentially high-dimensional variable sets. While CCA is concerned with revealing multivariate sources of variation based on linear covariance structure, ICA can identify more complicated non-linear relationships in data that can capture a scope of statistical relationships that go beyond differences in means and variance (65). A second aspect that departs from CCA is the fact that latent dimensions obtained from ICA are not naturally ordered from highest to lowest explained variance, which would need to be computed in a later step. Yet another key distinction lies in CCA's orthogonality constraint to obtain *uncorrelated* latent dimensions; in contrast, ICA optimizes the *independence* between the emerging hidden sources of variability. Independence between two variables implies their uncorrelatedness, but the lack of a *linear* correlation between the two variables does not ensure the lack of a *nonlinear* statistical relation between the two variables. Finally, it is worth mentioning that ICA can also be used as a post-processing step to further inspect CCA solutions (5,66).

iii) **Partial least squares (PLS) regression** is more similar to CCA than PCA or ICA. This is because PLS and CCA can identify latent dimensions of variability across *two* variable sets (67). While PLS is consistently viewed and used as a supervised method, it has been controversial whether CCA should counted as part of the supervised or unsupervised family (23). Further, the two methods are similar in the sense that they impose an orthogonality constraint on the hidden sources of variability to be discovered. However, the two methods are different in the optimization objective: PLS maximizes the variance and correlation of the projected dimensions with the original variables of the associated response. Whereas CCA operates only in embedding spaces of the left and right variable sets to maximize the correlation between the emerging low-rank projections, and thus indirectly identifies those canonical vectors whose ensuing canonical variates correlate most. In contrast to CCA, PLS is scale-variant by reliance on the covariance, which leads to different results after transforming the variables.)

Model extensions:

i. **Probabilistic CCA** is a modification that motivates classical CCA as a generative model (49,68). One advantage of this CCA variant can be seen in the more principled definition of variation to be expected in the data and the added possibility to produce synthetic but plausible observations once the model has been fit. Additionally, by virtue of this introduction of prior knowledge into the model specification, an advantageous aspect of many Bayesian models, these generative CCA approaches have been shown to yield more convincing results in small biomedical datasets which would otherwise be challenging to handle using ordinary CCA (e.g. ,69,70).

ii. **Sparse CCA** (SCCA ,21) is a variant for identifying parsimonious sources of variation by encouraging exactly-zero contributions from many variables in each variable set.

Besides facilitating interpretation of CCA solutions, the imposed $\ell_1$-norm penalty term is also effective in scaling CCA applications to higher-dimensional variable sets, where the number of variables can exceed the number of available observations (58). It can be both an advantage or disadvantage that the sparsity constraint interferes with the orthogonality constraint of CCA. Practically, the sparser the CCA modes, the more the canonical variates of the different modes can be correlated with one another other. Additionally, it is important to note that the variance that each mode explains will not decrease in order in SCCA, in contrast to ordinary CCA. As a side node, other regularization schemes can be interesting to extent classical CCA. In particular, imposing an $\ell_2$-norm penalty term stabilizes CCA estimation in the wide-data setting using variable shrinkage, without the variable-selection property of the sparsity-inducing constraint (71).

iii. A principled extension of CCA to also capture nonlinear relationships has been introduced by **kernel CCA** (KCCA; ,72). Kernels are mapping functions that implicitly express the variable sets in richer feature spaces, without ever having to explicitly compute this mapping, a method known as the "kernel trick" (23). KCCA first projects the data into this enriched feature space before performing CCA in that enriched input space. It is advantageous that KCCA allows for the detection of complicated non-linear relationships in the data. The drawback is that the interpretation of variable contributions in the original variable space is typically more challenging and in certain cases impossible. Further, KCCA is a nonparametric method; hence, the quality of the model fit scales poorly with the size of the training set (11).

iv. With the staggering advances in "deep" neural-network algorithms (73,74), **deep CCA** (DCCA ,75) has been introduced to further expand the representational capacity of KCCA. A core property of many modern neural network architectures is the capacity to learn representations in the data that emerge through multiple nested non-linear transformations. By analogy, DCCA simultaneously learns two deep neural network mappings of the two variable sets to maximize the correlation of their (potentially highly abstract) latent dimensions, which may remain opaque to human intuition.

## 5 PRACTICAL CONSIDERATIONS

After these conceptual considerations, we now provide several pointers for the practical application of CCA in everyday research. The computation of CCA solutions is made straightforward by built-in libraries in MATLAB (canocorr), R (cancor or the PMA package), and the Python machine-learning library scikit-learn (sklearn.cross_decomposition.CCA). The sparse CCA mentioned in the examples is implemented in R package PMA. These code implementations provide comprehensive documentation for how to deploy CCA. Here, we point out additional steps that investigators may want to keep in mind when applying CCA to their data.

## 5.1 Preprocessing

Some minimal data preprocessing is usually required as for most machine-learning methods. CCA is *scale-invariant* in that applying some standardizing data transformation on the columns of the variable sets should not change the resulting canonical correlations. This property is inherited from Pearson's correlation defined by the degree of simultaneous unit change between two variables, with implicit standardization of the data. Nevertheless, z-scoring of each variable of the measurement sets is still recommended before performing CCA to facilitate the model estimation process and to enhances interpretability. To avoid outliers skewing CCA estimation, it is recommended that one applies outlier detection and other common data-cleaning techniques (61). Several readily applicable heuristics exist to identify unlikely variable values, such as replacing extreme values with $5^{th}$ and $95^{th}$ percentiles of the respective input dimension, a statistical transformation known as winsorizing. Missing data is a common occurrence in large dataset. It is recommended to exclude observations with too many missing variables (e.g. those missing a whole domain of a questionnaire). Alternatively, missing variables can be "filled in" with mean or median when the proportion of missing data is small.

Besides unwarranted extreme and missing values, it is often necessary to account for potential nuisance influences on the variable sets. Deconfounding procedures are a preprocessing step in most analysis settings to reduce the risk of finding non-meaningful modes of variation. The same procedures that are also commonly applied prior to the use of linear-regression analyses can also be used in the context of CCA. Note that deconfounding is typically performed as an independent preceding step because the CCA model itself has no noise component. Deconfounding is done by creating a regression model that estimates variance of original data explained by the confounder. The residuals of such regression model will be the data with potential confound information removed. In neuroimaging, for example, head motion, age, sex, and total brain volume have frequently been considered unwanted sources of influence in many analysis contexts (5,15,42,76,77). While some previous studies have submit one variable set to a nuisance-removal procedure, in the majority of the analysis scenarios the identical deconfounding step should probably be applied on each of the variable sets.

## 5.2 Data reduction

When the number of variables exceeds the number of samples, dimensionality-reduction techniques can provide useful data preprocessing before performing CCA. The main techniques includes features selection based on statistical dispersion, such as mean or median absolute deviation, and matrix factorizing methods, such as PCA and ICA. As a common choice, the application of PCA compresses the number of variables in each matrix to a small set of most explanatory dimension of variation. The benefit of PCA dimension reduction is that CCA can subsequently be performed on a smaller, computationally more feasible set of variables. To interpret the CCA solutions in the original data, these authors have related the canonical variates with the original data to recover the relevant variate relationships with the original variables as captured by each CCA mode. A potential limitation of performing the PCA first before CCA is that the assumptions implicit in the PCA application carry over into the CCA solution.

Another attractive analysis strategy is to *post*-process the CCA solution via ICA (5,66). Such analysis tactic can overcome some issues of back-projecting the PCA-compressed data back into the original variable space. After CCA has been fitted, the ensuing canonical variates of both the left and right side can be concatenated across participants into one array (number of observations x 2 * number of modes). ICA is then applied to the aggregated canonical mode expressions to recover the *independent* sources of the variation *between observations expressed in the embedding space*. While incurring additional computational load, this approach can be advantageous because CCA can only disentangle latent directions of variation in the data up to a random rotation. The latent dimensions described by the obtained canonical vectors and variates can be further disambiguated by the final ICA step (66). Going beyond discovery of *uncorrelated* sources of variation, the ICA post-processing is especially useful in the detection of *independent* components that contribute to the common solution extracted from the two variable sets. The CCA+ICA approach could zoom in more on relationships between the two original variable sets in some cases. Yet, additional application of PCA preprocessing could influence the outcome of the CCA+ICA approach (66).

## 5.3 MODEL SELECTION

CCA allows multiple modes to be calculated from the observed data. It is frequent question how to choose the optimal number of latent sources of variation to be extracted. While various strategies have been proposed, little consensus has been achieved so far. The current ambiguity around how to choose this CCA hyper-parameter is closely related to issues of choosing the number of clusters in k-means and other clustering procedures as well as choosing the number of components in PCA, ICA, and other matrix decomposition techniques (78).

To select a useful number of modes, several quality metrics can be used for quantifying the variance explained to operationalize a notion of the optimal sources of variation, without a clear default. Since the canonical variates represent the compressed (i.e., projected) information of the original data, the canonical modes should bear relation to the original data. Among other alternatives, one useful strategy is to assess the decrease in that explained variance metric with the canonical variate of one domain to predict the original variables with increasing number of modes (33). A drop in the variance captured after adding yet another mode for modeling k+1 sources of variability indicates a candidate cut-off at k. An important overarching property in this context is that computing classical CCA with, say, 5 modes and then, say, 50 modes produces the identical first 5 modes (i.e., related to the orthogonality constraint).

As such, another tactic relies on determining how many of the extracted modes are statistically robust as indicated by non-parametric permutation tests (e.g. ,15,42). An empirical distribution of canonical correlation of each mode can be computed under the null hypothesis that there is no coherent relation between the left and right variable set - the canonical correlation should hence fluctuate around chance level. The permutation procedure proceeds by random shuffling of the rows or columns of the two variable sets to break any existing relationships between the ensuing low-rank projections of the two variable sets across observations (79,80). If the relation between the two variable sets is random, all derived modes should be meaningless. The first mode can be viewed as the strictest measure of null hypothesis, because it extracts the highest explained variance in a null sample (e.g. ,15). In typically hundreds of such permutation iterations of repeated CCA application, the extracted perturbed mode from the permutation datasets serve to compute the chance level of associations between the two variable sets. Each canonical mode whose original canonical correlation exceeds the 95% level (significance at $p < 0.05$) or 99.9% level (significance at $p < 0.001$) can be certified as robust under the null hypothesis of absent linkage between the left and right variable set. If the investigator wishes to add an explicit correction for multiple comparisons, the p-value threshold can for instance be divided by the number of modes (i.e., Bonferroni's method) or false-discovery rate (FDR) can be used to reduce possible type I errors. This approach hence yields one p-value for each of the originally obtained CCA modes. Please note that no statistical null hypothesis testing is performed on an individual variables in this way, which illustrates again CCA's weakness to make targeted statements about specific input variables.

Moreover, a hold-out framework has been proposed to determine the generalizability and statistical significance of discovered CCA modes in sufficiently large samples (81,82). This analysis scheme starts by randomly separating the data into a training set and a holdout set. A CCA model is then fitted from the training set. The data from held-out individuals is then projected to the previously obtained CCA embedding (i.e., using the precomputed canonical vectors) to generate independent hold-out correlations. Then, a permutation test is done on the test data against the left-out correlations. This validation framework can be used to explicitly measure the pattern-generalization performance and obtain a p-value for the mode. A possible limitation lies in the required reasonably sized hold-out set.

Finally, to explicitly evaluate the contribution of each individual input variable to the overall modeling solution, a sensitivity analysis has been performed for CCA (42). The impact of each variable was isolated by selectively removing all information from a given input variable, including for instance the functional connectivity strengths derived from that same brain region, and reiterating the CCA procedure based on the reduced data of one variable set and the original data from the other variable set. This analysis strategy issued a perturbed set of canonical variates under the assumption that, one-by-one, a particular input dimension may not have been important to obtain the original canonical modes. The degree of alteration in the canonical correlations was quantified by computing Pearson's correlation coefficient between the original and perturbed canonical variates. In addition to these point estimates after variable deletion, the induced statistical uncertainty was quantified by carrying out bootstrapping analysis. "Shaken up" bootstrap datasets were generated from the original participant sample by randomly drawing individuals with replacement. In each of these alternative datasets, the perturbed CCA was fitted and evaluated in identical fashion. This robustness assessment provided population-level uncertainty intervals and hence enabled extrapolation of statements on variable importance on data we would observe in the future. High correlation between the original canonical variates and the canonical variates obtained without the contribution of a specific variable exhibited indicated that the variable in question was not vital to estimating the original CCA correspondence between the two data modalities. This is because removing the give variable (and any related information) incurred no dramatic change of the original CCA performance metrics. Instead, low correlations pointed towards variables that were of special relevance for deriving the covariation between the two levels of observations. This generally applicable variable-deletion scheme can determine interpretable contributions of single input variables that play disproportionately important roles in highly multivariate analyses such as CCA.

# 6 SUMMARY

In much biomedical research, the relationships among body, brain, cognition, and genes, and their impact on disease are assumed to be complicated. Satisfying growing ambitions towards "deep" phenotyping and modality fusion, CCA provides a clean methodology to conduct a pattern search for uncovering essential correspondences between two rich variable sets. The appeal of CCA is likely to increase as the detail and quality of multi-modal datasets is now growing rapidly in neuroscience and medicine (11,60).